%% file: main.tex
\documentclass{article}
\usepackage{spconf,amsmath,amssymb,graphicx,booktabs}
\usepackage[hidelinks]{hyperref}
\makeatletter
\renewcommand\@makefntext[1]{\noindent#1}
\makeatother
\title{ACTION-CONDITIONED BISIMULATION FOR GUI AGENT MEMORY}
\name{Hongbo Zhang$^{1,*}$, Liuyang Song$^{1,*}$, Quanquan Li$^{2}$,
Daqian Yang$^{1}$, Yan Wen$^{3}$, Zhengtao Yao$^{3,\dagger}$
\thanks{\raggedright $^{*}$Equal contribution.\\
$^{\dagger}$Corresponding author: \texttt{zyao9248@alumni.usc.edu}.}}
\address{$^{1}$Peking University \quad $^{2}$East China Normal University \quad
$^{3}$University of Southern California}
\begin{document}
\ninept
\maketitle
\begin{abstract}
An agent that remembers what it did on a web page must decide when two pages
count as the same. Memories built on observation similarity merge pages
that look alike but behave differently, and GUIs are full of such pages:
two tabs of one widget or two rows of one menu answer the same click
differently. We define the merge rule as an action-conditioned bisimulation over
the empirical predictive state graph a frozen agent fills as it acts. Two states
merge only when their shared actions lead to agreeing outcomes and successor
blocks under an affordance label. Observation similarity never enters the rule,
and nothing is trained. It replaces the merge rule of an existing
outcome-value memory, so a closed-loop comparison isolates it. On MiniWoB++ it
raises success rate over a memoryless agent, while a control
taking identical exploratory detours, the prior successor-representation merge,
and the same criterion without action conditioning change nothing. 
\end{abstract}
\begin{keywords}
GUI agents, state abstraction, bisimulation, agent memory
\end{keywords}

\section{Introduction}
A GUI agent that is not retrained between episodes can still improve, but only
if it can recognise that it has been here before. Every memory for such an agent
therefore rests on one decision. Given two observations, are they the same
state? Get it wrong in one direction and each episode is a fresh graph, so
nothing transfers; get it wrong in the other and experience from one page is
recalled on a different page that merely resembles it.

The prevailing answer is similarity. Synapse \cite{synapse} retrieves abstracted
trajectories by embedding distance, on the benchmark used here; ExpeL
\cite{expel} keeps successful trajectories beside insights distilled from them;
Agent Workflow Memory \cite{awm} induces reusable sub-routines; AutoGuide
\cite{autoguide} recalls the guideline whose context matches the present state;
A-MEM \cite{amem} links notes into a network indexed by embedding; and
successor-representation memories \cite{dayan} merge states whose discounted
occupancy profiles agree. They differ in what they store and agree on how they
bring it back. An item returns when the present observation resembles the one it
was stored under. That answers \emph{which memory is relevant here}, not
\emph{is this the same state}.

On a GUI the two questions come apart adversarially. Two
tabs of the same widget, two rows of the same menu, and a control before and
after it has been enabled are near-identical as observations and differ entirely
in their response to the same click. Merging them pools evidence across two
different action semantics, and the agent then recalls a confident, wrong
action.

\begin{figure*}[t]
\centering
\includegraphics[width=\textwidth]{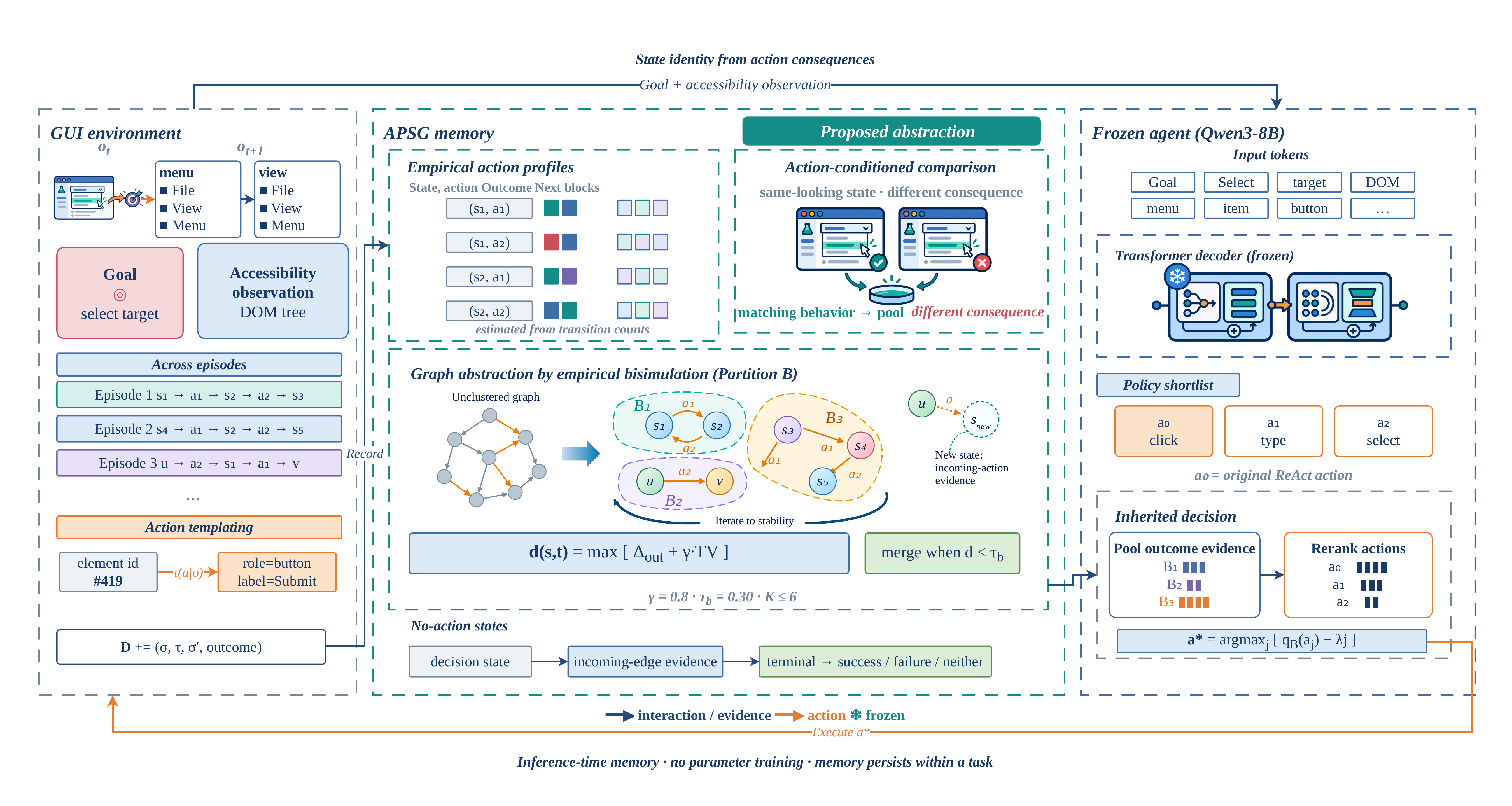}
\caption{The loop the criterion sits in. A frozen policy acts on the GUI and
every transition is recorded, so each state accumulates an empirical profile of
what its tried actions led to. States are compared only through those profiles,
never through what the pages look like, and iterating the comparison to
stability gives the partition $B$. Everything to the right of $B$ is inherited
unchanged from the outcome-value memory the criterion is dropped into.}
\label{fig:framework}
\end{figure*}

We replace the criterion with an action-conditioned predictive state graph
(APSG, Fig.~\ref{fig:framework}). A state is characterised by what it predicts
under each action, the immediate outcome and the distribution over successor
states, and two states merge only when those predictions agree for every action
both have actually tried. This is the bisimulation criterion \cite{givan,ferns}, which underpins representation
learning for deep RL \cite{zhang} and is still being revised
\cite{revbis,luo}. That line learns an embedding offline, from reward, in
continuous control. We use the criterion as what it originally was, a partition,
compute it by counting transitions a deployed agent has already made, and train
nothing.

Native GUI models such as UI-TARS \cite{uitars} improve the perception and
grounding of the policy itself, and AgentOccam \cite{agentoccam} shows on
WebArena \cite{webarena} that aligning the observation and action space with the
policy outperforms added orchestration. We take the same view from the memory
side.

\section{Method}

\subsection{The empirical transition graph}
\label{sec:actions}
A frozen agent interacting with a GUI produces transitions $(s,a,s')$, in which
$s$ indexes an observation signature $\sigma(o)$ and $a$ an action template
$\tau(a\mid o)$. A template replaces the transient parts of an action
with descriptors that survive the episode, an element id by the element's
accessibility role and label and a goal-supplied label by the slot it filled,
so that clicking \texttt{Donetta} in one episode and \texttt{Chery} in the next
yields one template. It resolves backwards against the current page, failing
closed unless each descriptor matches one element. The trajectories $o_0,a_0,o_1,\dots$ of every episode of a task then accumulate into
the transition multiset
\begin{equation}
 \mathcal{D}=\bigl\{\bigl(\sigma(o_i),\ \tau(a_i\mid o_i),\
 \sigma(o_{i+1})\bigr)\bigr\}_{i},
\label{eq:data}
\end{equation}
whose distinct first coordinates are the states $S$. With $n(s,a,s')$ counting
the copies of $(s,a,s')$, $n(s,a)=\sum_{s'}n(s,a,s')$, $A(s)=\{a:n(s,a)>0\}$ the
actions ever tried at $s$, and $S^{\pm}$ the states at which an episode
terminated in success or failure, the entire description of a state used here is
its empirical prediction under each action tried there,
\begin{equation}
 P(s'\mid s,a)=\frac{n(s,a,s')}{n(s,a)},\qquad
 \rho^{\star}(s,a)=\!\!\sum_{s'\in S^{\star}(s)}\!\! P(s'\mid s,a),
\label{eq:rho}
\end{equation}
with $S^{\pm}(s)=S^{\pm}$ and the progress successors $S^{\triangle}(s)$ defined
next.

Progress needs care, since tabs and menus repaint a page in both directions and
reading $\rho^{\triangle}$ as ``the page changed'' would credit oscillation.
Writing $u\rightsquigarrow v$ for reachability over the edges with $n(u,a,v)>0$,
we grade a transition by $c(s,s')$, which is $-1$ on a self-loop, $0$ when
$s'\rightsquigarrow s$ and $1$ otherwise. Progress is then only what has no
recorded way back, $S^{\triangle}(s)=\{s':c(s,s')=1\}$.

\subsection{Action-conditioned bisimulation distance}
\begin{figure}[t]
\centering
\begin{minipage}{0.315\linewidth}\centering
 \includegraphics[width=\linewidth]{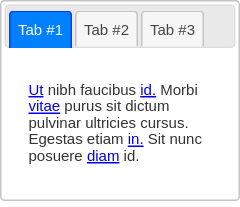}\\[2pt]
 {\scriptsize (a) Tab \#1\\ \texttt{click(18)}}
\end{minipage}\hfill
\begin{minipage}{0.315\linewidth}\centering
 \includegraphics[width=\linewidth]{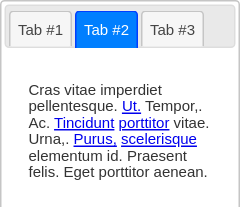}\\[2pt]
 {\scriptsize (b) Tab \#2\\ \texttt{click(20)}}
\end{minipage}\hfill
\begin{minipage}{0.315\linewidth}\centering
 \includegraphics[width=\linewidth]{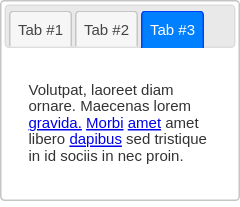}\\[2pt]
 {\scriptsize (c) Tab \#3\\ \texttt{click(22)}}
\end{minipage}

\vspace{6pt}
{\scriptsize\setlength{\tabcolsep}{2.8pt}
\begin{tabular}{l|cccccccc|l}
\toprule
step & 1 & 2 & 3 & 4 & 5 & 6 & 7 & 8 & \\
\midrule
ReAct & 20 & 18 & 20 & 18 & 20 & 18 & 20 & 18 & \emph{fail} \\
APSG & 20 & 18 & 20 & 18 & 20 & \textbf{22} & 40 & --- & \emph{success} \\
\bottomrule
\end{tabular}}
\caption{One evaluation cell of \texttt{click-tab-2} (goal
``Switch between the tabs to find and click on the link \texttt{gravida.}'').
The panels are the three states reachable before the target is visible; only
(c) holds the link. They differ by a highlight and a paragraph of filler, so an
observation-similarity criterion merges them, while their responses to the same
three clicks differ and \eqref{eq:dist} separates them. Below, the element
clicked at each step. ReAct alternates Tab \#2 and Tab \#1 for all eight steps.
APSG emits the same five actions, then overrides at step $6$ (bold), once the
block holds enough recorded failures of \texttt{click(18)} for
\eqref{eq:rerank} to prefer \texttt{click(22)}.}
\label{fig:cell}
\end{figure}

Let $B$ be a partition of the states, $B(s')$ the block of $s'$, and $P_B$ the
pushforward of $P$ onto blocks. For an action $a$ tried at both $s$ and $t$,
\begin{equation}
\begin{split}
 d_a(s,t)=\ &\underbrace{\tfrac{1}{3}\!\!\sum_{\star\in\{+,-,\triangle\}}\!\!
 \bigl|\rho^{\star}(s,a)-\rho^{\star}(t,a)\bigr|}_{\text{immediate outcome}}\\[2pt]
 &+\ \gamma\,\underbrace{\mathrm{TV}\bigl(P_B(\cdot\mid s,a),
 P_B(\cdot\mid t,a)\bigr)}_{\text{successor blocks}},
\end{split}
\label{eq:gap}
\end{equation}
with $\mathrm{TV}$ the total variation distance. The state distance is the worst
case over the actions that both states have tried, and is infinite
unless they carry the same label $\ell$,
\begin{equation}
 d(s,t)=\begin{cases}
 \infty & \text{if }\ell(s)\neq\ell(t)\\
 & \text{or }A(s)\cap A(t)=\emptyset,\\[2pt]
 \displaystyle\max_{a\in A(s)\cap A(t)} d_a(s,t) & \text{otherwise.}
 \end{cases}
\label{eq:dist}
\end{equation}
The restriction to shared actions is the substantive modelling choice. An action
observed at $s$ alone carries no evidence about $t$, and treating its absence as
agreement would merge on missing data, which is how a similarity rule fails.
States with no evidence in common are never merged, however alike they look, and
the trio of Fig.~\ref{fig:cell} shows the criterion separating look-alikes that
do share actions. It also makes the
action vocabulary load-bearing. A state carries only the few actions tried
there, often one, so under actions named by the element id clicked every
cross-episode pair is at infinite distance and the abstraction degrades to the
identity, which the templates of Sec.~\ref{sec:actions} prevent.

The label is what bisimulation on a labelled system requires before it compares
transitions at all \cite{givan}, and the action-conditioned term refines it
rather than replacing it. We take $\ell(s)$ to be the invariant affordance
signature of the page,
\begin{equation}
 \ell(s)=\bigl(\hat{g},\ \{\!\{\mathrm{role}(e):e\in E(s)\}\!\},\ \pi(s)\bigr),
\label{eq:label}
\end{equation}
in which $\hat{g}$ is the goal with each named target replaced by its slot
position, $E(s)$ the elements of the page, and $\pi(s)$ the goal slots fillable
on it, all properties of the task and not of the episode. It tests what a page
offers rather than what it displays, and is inherited from the memory we build
on (Sec.~\ref{sec:inherited}). Dropping it is not harmless. A closed menu and
the menu it opens into afford the same toggle and are each other's successor
under it, so both terms of \eqref{eq:gap} agree once the pair is merged, and the
refinement settles on a coarse fixed point in which the pooled value of the
toggle outranks the action that finishes the task.

\subsection{The partition is a fixed point}
Equation~\eqref{eq:gap} compares distributions over blocks of $B$, so $B$ appears
on both sides of its own definition and we compute it as a fixed point. Writing
$d^{B}$ for \eqref{eq:dist} evaluated with the pushforward onto $B$ and
$\mathrm{UF}(B,\mathcal{P})$ for the union--find closure of $B$ under the pairs
$\mathcal{P}$ in increasing order of $d^{B}$, we start from the immediate-outcome
partition $B_0$, which is \eqref{eq:dist} at $\gamma{=}0$, and iterate
\begin{equation}
 B_{k}=\mathrm{UF}\bigl(B_{k-1},\ \{(s,t)\ :\ d^{B_{k-1}}(s,t)\le\tau_b\}\bigr)
\label{eq:fixed}
\end{equation}
until $B_k=B_{k-1}$ or a budget $K$ is spent; since \eqref{eq:fixed} only
coarsens, the iteration is monotone and settles after a few rounds in practice.
This is the usual bisimulation-metric construction \cite{ferns,castro}, run on
empirical rather than known transitions. The graph of
Sec.~\ref{sec:actions} carrying this partition is the APSG.

A state with $A(s)=\emptyset$ supports no comparison under \eqref{eq:dist},
which is no corner case, since it covers every terminal and the state currently
being decided. Left a singleton it returns nothing from \eqref{eq:q}, so we place it instead by
its incoming signature, which is action-conditioned in the same sense as
\eqref{eq:dist}. Given a recorded $(u,a,s)$, it joins the block that transitions
leaving $B(u)$ under $a$ most often landed in; an episode's first page joins the
block its counterparts in past episodes occupy, and only a terminal falls back
to its role as success, failure, or neither. Each placement obeys the label
requirement.

\subsection{What is deliberately inherited}
\label{sec:inherited}
The abstraction is the whole method, so everything downstream is held fixed. We
implement the rule as a subclass of an existing outcome-value memory that
overrides the partition and nothing else. That memory takes
$V=(1-\gamma)(I-\gamma P_{\mathcal{D}})^{-1}r$ with
$r=\mathbf{1}_{S^{+}}-\mathbf{1}_{S^{-}}$, the normalised discounted terminal
occupancy of the recorded process, and scores a transition by the empirical
advantage $\delta(u,v)=g(v)-V(u)$, whose continuation $g(v)$ is $\pm1$ at a
terminal in $S^{\pm}$ and $\gamma V(v)$ otherwise. The abstraction enters only
now. The evidence for a template $a$ at the current state $s$ is pooled over the
whole block $B(s)$,
\begin{equation}
 q_B(a)=\frac{\displaystyle\sum_{(u,a,v):\,B(u)=B(s)}\!\!\!
 n(u,a,v)\,\bigl[\delta(u,v)+\eta\,\bar c(u,v)\bigr]}
 {\displaystyle\kappa+\!\!\!\sum_{(u,a,v):\,B(u)=B(s)}\!\!\! n(u,a,v)},
\label{eq:q}
\end{equation}
with $\eta\,\bar c$ a small dense term ($\bar c=c$, but $-1$ at a recorded
failure) and $\kappa$ a prior against one transition dominating. Given the policy's own shortlist $a_0,\dots,a_{m-1}$ in
its own order, the memory returns
\begin{equation}
 a^{\star}=\arg\max_{0\le j<m}\ q_B(a_j)-\lambda j,
\label{eq:rerank}
\end{equation}
with $q_B(a_j)=0$ where no evidence resolves onto $a_j$, so that the rank penalty
$\lambda$ makes an override require the recalled evidence to outweigh the
policy's own ordering and $a_0$ survives whenever the memory is silent.

All of this is inherited verbatim, including the label of \eqref{eq:label},
which that memory already gates its own merges on, so that the change of
criterion is not confounded with a relaxation of the label. The three merge
rules therefore differ in the single symbol $B$ of \eqref{eq:q}, and nothing
else is added, no router, no certificate, no stagnation heuristic, and no branch
on the task identity.

\subsection{Making \texorpdfstring{$B$}{B} observable}
\label{sec:observable}
Two properties of a deployed greedy agent would make any comparison of merge
rules vacuous. First, \eqref{eq:rerank} needs candidates, which sampling cannot
supply once thinking is disabled, so we keep the policy's greedy action as the
head $a_0$ and append distinct alternatives from a second greedy call; a memory
that never overrides then reproduces the baseline exactly. Second, greedy
execution leaves nothing to pool, since the executed action is a function of the
page, so $|A(s)|=1$ however often $s$ is visited and $q_B$ can only re-propose
the one action it ever saw, whatever $B$ is. The reranking systems therefore
take an $\epsilon$-greedy detour off $a_0$ over the first third of each run,
seeded by the environment seed and step index and never by the system, so a
memoryless control makes the identical detours and only what is kept of them
differs.

\section{Experiments}
\label{sec:results}

\input{tables_apsg}

\begin{figure}[t]
\centering
\includegraphics[width=0.93\columnwidth]{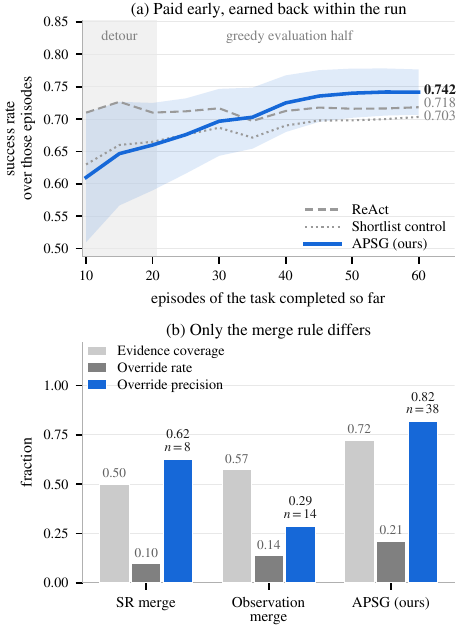}
\caption{Both panels recomputed from the episode log. \emph{(a)} Cumulative
success rate, with a 95\% bootstrap band on our curve only. The shortlist
control takes the identical detours but keeps nothing, so it recovers to
ReAct's rate and stops there; under the action-conditioned partition the same
detours are paid off by episode $35$. \emph{(b)} On the evaluation episodes:
\emph{evidence coverage}, the share of decisions whose block held recorded
evidence for a candidate; \emph{override rate}, the share at which
\eqref{eq:rerank} displaced the policy's first choice; \emph{override
precision}, the share of displaced outcomes that moved from failure to success
rather than the reverse.}
\label{fig:analysis}
\end{figure}

We evaluate closed-loop on MiniWoB++ \cite{miniwob} through BrowserGym
\cite{browsergym}, on ten navigation-style tasks fixed beforehand in which the
same page layouts recur across episodes while each episode draws a different
environment instance, so cross-episode state identity decides the outcome. The
policy is Qwen3-8B \cite{qwen3} served locally, frozen, thinking
disabled, and capped at eight steps. Each task runs sixty episodes; episodes
$1$--$20$ take the detours of
Sec.~\ref{sec:observable} and episodes $21$--$60$ are greedy for every system.
One setting serves every task and system ($\epsilon=0.35$, $\tau_b=0.30$,
$\gamma=0.8$, $K=6$, and the inherited $\lambda=0.015$, $\kappa=2$,
$\eta=0.10$). A hand-written shortcut sits in front of the policy in all five
systems, acting without a model call whenever exactly one visible element
carries the label the goal names; it fired $171$ times for ReAct \cite{react}
and $170$ times for each of the others, so it cannot separate them.

\textbf{The ablation ladder.} Table~\ref{tab:main} is read from the bottom
up. Each rung holds the policy, the prompt, the model-call budget and the
exploration draws fixed and changes exactly one thing, and three of the four
rungs fail to move. The shortlist control pays the memories' second model call
and takes every one of their detours, yet ends the evaluation episodes without a
single discordant pair against ReAct, so neither the extra call nor the
exploration finds better actions by itself. Keeping that same evidence under the
SR merge is indistinguishable from keeping none of it, and the observation
merge is if anything worse than not merging at all. Only the last rung moves, by $+0.060$ against ReAct
($p<10^{-3}$) and by $+0.075$ against the observation merge it replaces. What
separates the rules is which states they merge, not how many. The observation
merge is the more cautious, collapsing the graph roughly twofold against our
roughly fivefold, and it is the one that loses, since pooling evidence helps
only when the states pooled answer the same action the same way.
Fig.~\ref{fig:analysis}(b) locates the difference at the moment of decision.
Neither rule is silent, so the criterion does not make the memory quieter; it
makes it right more often when it does speak. Ours moves most of the outcomes
it touches from failure to success and the observation merge moves most of them
the other way, which is the aliasing of Fig.~\ref{fig:cell} showing up at the
decision itself. A confident wrong recall costs more than no recall, and the
shared-action requirement of \eqref{eq:dist} is the clause that refuses the
pair that produces one.

\textbf{Where the gain falls and what it costs.} Before the run we split the
suite by a stated property, not by outcome, and recorded that the gain should
concentrate on one side and be absent on the other. A container task is one
whose goal names one of several interchangeable containers (tabs, menus,
disclosure triangles), so that pages differing in identity coincide in
behaviour. Both halves of that prediction hold. On the container tasks the
difference against ReAct is $+0.075$, and on the others there is no discordant
pair for any memory system on any cell: where there is nothing to alias,
\eqref{eq:rerank} returns the policy's own action because no block has
accumulated evidence against it, so those tasks are a negative control the
mechanism passes, not a subset it is weak on. Inside the container set the
movement concentrates on the tab and collapsible tasks, where near-identical
pages differ only in what they open, while the tasks already at ceiling show
the memory doing no harm. One container task,
\texttt{click-menu-2}, moves against us; its cause remains unresolved. The gain
is bought with exploration, and Fig.~\ref{fig:analysis}(a) shows the cost
repaid. The memory starts behind, having spent a third of the run deliberately
off-policy, crosses ReAct within the run, and ends ahead of it over every
episode, not only the evaluation ones, while the control pays the
identical cost and recovers to ReAct's rate and no further. On the $272$ cells every system solves, where only
directness can differ, it reaches the goal in fewer steps than any other system
($1.70$ against ReAct's $1.81$), because a filled block records which of the
look-alike pages is worth opening.

\textbf{What a merge rule trades, and why this one bounds it.} Everything the
memory can do passes through \eqref{eq:q}, a mean of empirical advantages taken
over a block, so a merge rule has one way to help and one way to hurt. It helps
by count. A single page is visited a handful of times, $\kappa$ dominates its
denominator, and an unmerged memory is therefore silent rather than wrong; a
block puts enough recorded transitions behind a template for the numerator to
outweigh the prior, which is the coverage every rule in
Fig.~\ref{fig:analysis}(b) buys. It hurts by bias. The pooled mean stands in
for the advantage of $a$ at $s$ only insofar as the states merged with $s$
answer $a$ as $s$ does, and any disagreement left among them enters $q_B(a)$
carrying the full weight of their counts, with nothing downstream able to
detect it. The two move together under any rule, and the criterion is the only
place the second is bounded. Equation~\eqref{eq:gap} is built from the two ways
the summand of \eqref{eq:q} can move between $s$ and $t$ under one action. Its
advantage $\delta(u,v)=g(v)-V(u)$ separates into the outcome recorded at the
transition, which the first term compares directly, and a continuation
$\gamma V(v)$, which depends on $v$ through its block alone exactly when $B$ is
a bisimulation of the recorded process, and which the second term therefore
compares as a distribution over blocks. Thresholding $d_a\le\tau_b$ caps the
disagreement a merge admits, the maximum in \eqref{eq:dist} makes the cap a
worst case over the shared actions and not an average, and refusing $A(s)\cap
A(t)=\emptyset$ keeps it from being a maximum over an empty set. The cap is
stated on the partition it is used to build, which is why \eqref{eq:fixed}
iterates: each round re-evaluates every admitted merge against the coarser
blocks its own predecessors created, so the property the bound assumes is the
property the fixed point establishes. Merging on appearance buys the same
counts with no such object to iterate towards, and the bias it admits is
invisible to the rule that admitted it.

\clearpage
\sloppy

\end{document}

%% file: tables_apsg.tex

\begin{table}[t]
\caption{Closed-loop MiniWoB++ over 10 tasks with a frozen Qwen3-8B, single seed. SR is the benchmark's own success rate on the evaluation half fixed before the run (400 paired cells per system), decoded greedily by every system. Environment seeds are identical across systems, so $\Delta$ against ReAct is paired cell by cell, with a 95\% percentile bootstrap interval and an exact McNemar $p$.}
\label{tab:main}
\centering\scriptsize\setlength{\tabcolsep}{3pt}
\begin{tabular}{lcccc}
\toprule
System & SR & $\Delta$ vs.\ ReAct & 95\% CI & $p$ \\
\midrule
\multicolumn{5}{l}{\emph{Memoryless}} \\
\quad ReAct & 0.723 & -- & -- & -- \\
\quad Shortlist control & 0.723 & $+0.000$ & $[+0.000,+0.000]$ & 1.000 \\
\midrule
\multicolumn{5}{l}{\emph{Memory, prior merge rules}} \\
\quad SR merge & 0.728 & $+0.005$ & $[-0.007,+0.018]$ & 0.727 \\
\quad Observation merge & 0.708 & $-0.015$ & $[-0.033,+0.003]$ & 0.180 \\
\midrule
\multicolumn{5}{l}{\emph{Memory, ours}} \\
\quad \textbf{APSG (ours)} & \textbf{0.782} & $+0.060$ & $[+0.033,+0.090]$ & $<$0.001 \\
\bottomrule
\end{tabular}
\end{table}
